\documentclass{ifacconf}

\usepackage{graphicx}     
\usepackage{natbib}        % required for bibliography
\usepackage{xcolor}
\usepackage{amsmath}
\usepackage{amsfonts}
\usepackage{amssymb}
\usepackage{mathtools}
\usepackage{algorithmicx}
\usepackage{algorithm}
\usepackage{algpseudocode}

\usepackage{todonotes} % for comments

\newcommand{\displaycomments}
\ifdefined\displaycomments

\newif\ifAddInBlue
\newcommand{\add}[1]{%
  \ifAddInBlue
  \textcolor{blue}{#1}
  \else
    \textcolor{black}{#1}	
  \fi%
}
\newif\ifShowDel
\begin{document}
\begin{frontmatter}

% \title{Learning Parametric Differentiable Predictive Control Policies subject to Nonlinear Chance Constraints} 

\title{Learning Stochastic Parametric Differentiable Predictive Control Policies} 

% \title{Learning Stochastic Parametric Differentiable Predictive Control Policies subject to Nonlinear Chance Constraints} 
% \title{Learning Robust Neural Differentiable Predictive Control Policies} 

% \thanks[footnoteinfo]{Sponsor and financial support acknowledgment
% goes here. Paper titles should be written in uppercase and lowercase
% letters, not all uppercase.}

\author[First]{J\'an Drgo\v na} 
\author[First]{Sayak Mukherjee}
\author[First]{Aaron Tuor}
\author[First]{Mahantesh Halappanavar}
\author[First]{Draguna Vrabie}

\address[First]{Pacific Northwest National Laboratory,
	Richland, Washington, USA \\ e-mails: 	\{jan.drgona,
	sayak.mukherjee, aaron.tuor, mahantesh.halappanavar, draguna.vrabie\}@pnnl.gov}

\begin{abstract}                % Abstract of not more than 250 words.

The problem of synthesizing stochastic explicit model predictive control policies is known to be quickly intractable even for systems of modest complexity when using classical control-theoretic methods.
To address this challenge, we present a scalable alternative called stochastic parametric differentiable predictive control (SP-DPC) for unsupervised learning of neural control policies governing stochastic linear systems subject to nonlinear chance constraints.
SP-DPC is formulated as a deterministic approximation to the stochastic parametric constrained optimal control problem.
This formulation allows us to directly compute the policy gradients 
via automatic differentiation of the problem's value function, evaluated over sampled parameters and uncertainties.
In particular, the computed expectation of the SP-DPC problem's value function is backpropagated through the closed-loop system rollouts parametrized by a known nominal system dynamics model and neural control policy which allows for direct model-based policy optimization.
We provide theoretical probabilistic guarantees for policies learned via the SP-DPC method on closed-loop stability and chance constraints satisfaction.
Furthermore, we demonstrate the computational efficiency and scalability of the proposed policy optimization algorithm in three numerical examples, including systems with a large number of states or subject to nonlinear constraints.

\end{abstract}


\end{frontmatter}
%===============================================================================

\section{Introduction}

%\add{should we have some additional intro before related works?}
% \subsection{Related Work}

Model predictive control (MPC) approaches have been considered a viable modern control design method that can handle complex dynamic behavior and constraints. The presence of uncertainties in the system dynamics led to various forms of MPC designs robust to various non-idealities. Robust MPC (RMPC) approaches such as \citep{chisci2001systems} consider bounded disturbances and use constraint set tightening approaches to provide stability guarantees. \cite{pin2009robust} proposed a tube-based approach for nonlinear systems with state-dependent uncertainties. An event-triggered robust MPC approach has been discussed in \cite{liu2017aperiodic} for constrained continuous-time nonlinear systems. Robust MPC approaches, however, do not necessarily exploit the existing statistical properties of the uncertainty, thereby leading to conservative designs. Stochastic MPC (SMPC), on the other hand, accounts for some admissible levels of constraint violation by employing chance constraints. An overview of SMPC based approaches can be found in \cite{mesbah2016stochastic}. Works such as \cite{fleming2018stochastic,lorenzen2017stochastic,farina2016stochastic} developed methodologies that use a sufficient number of randomly generated samples to satisfy the chance constraints with probabilistic guarantees. 
Others~\citep{GOULART2006523,Primbs2009,Hokayem2009} have proposed feedback parametrization to cast the underlying stochastic optimal control problem tractable.
Some authors proposed approximation schemes that allow obtaining explicit solutions to stochastic MPC problems with linear~\citep{Drgona_SEMPC2013} or nonlinear~\citep{Grancharova2010} system models via parametric programming solvers. Unfortunately, these explicit stochastic MPC methods do not scale to systems with a large number of states and input of constraints.

Recently, more focus has been given toward Learning-based MPC (LBMPC) methods~\citep{ASWANI20131216}. These methods primarily learn the system dynamics model from data following the framework of classical adaptive MPC. 
% A recent review on the LBMPC approaches can be found in ~\cite{Hewing2020} and references therein. 
Combining LBMPC and RMPC approaches include formulation of robust MPC with state-dependent uncertainty for data-driven linear models~\citep{SOLOPERTO2018442}, iterative model updates for linear systems with bounded uncertainties and robustness guarantees~\citep{Bujarbaruah2018AdaptiveMF}, Gaussian process-based approximations for tractable MPC~\citep{hewing2019cautious}, or approximate model predictive control via supervised learning~\citep{DRGONA2018,LUCIA2018511}.

\cite{Zanon2019,ZanonRLMPC2019} use reinforcement learning (RL)  for tuning of the MPC  parameters.
 \cite{Kordabad2021} propose the use of scenario-tree RMPC whose parameters are tuned via RL algorithm.
%  While \cite{Lenz2015DeepMPCLD} discussed a recurrent neural model for learning latent dynamics for MPC. 
 In a learning-based MPC setting, various forms of probabilistic guarantees are discussed in recent works such as \cite{hertneck2018learning, rosolia2018stochastic, Karg2021}. 
 \add{
 Most recently, several authors~\cite{East2020,DiffMPC2018} have proposed differentiating the MPC problem for safe imitation learning.
Inspired by these directions, we proposed the use of   
 differentiable programming~\citep{DiffProg2019} for unsupervised
 deterministic  policy optimization algorithm called differentiable predictive control (DPC)~\cite{drgona2021learning,DRGONA202114}.
 Where DPC is based on casting the explicit MPC problem as a differentiable program implemented in a programming framework supporting AD for efficient computation of gradients of 
 the underlying constrained optimization problem.
 }
 
%  \subsection{Contributions}

\textit{Contributions.} 

\add{In this paper, we provide a computationally efficient, offline learning algorithm for obtaining parametric solutions to stochastic explicit MPC problems. 
The proposed method called stochastic parametric differentiable predictive control (SP-DPC) is an extension of the deterministic DPC~\cite{drgona2021learning,DRGONA202114}, with the main novelty being the inclusion of additive system uncertainties.
In particular, we leverage sampling of chance constraints in the stochastic DPC problem reformulation and introduce the stochastic model-based policy optimization algorithm. }

\add{
 From a theoretical perspective, we adapt probabilistic performance guarantees as introduced in~\cite{hertneck2018learning}
in the context of the proposed offline DPC  policy optimization algorithm.
However, contrary to the supervised learning-based approach  in~\cite{hertneck2018learning}, we propose an unsupervised  approach to solve the underlying stochastic parametric optimal control problem by leveraging differentiable programming~\citep{DiffProg2019}.}

 From an empirical perspective, we demonstrate the computational efficiency of the proposed SP-DPC policy optimization algorithm compared to implicit MPC via online solvers, with scalability beyond the limitations of explicit MPC via classical parametric programming solvers. In three numerical examples, we demonstrate the stochastic robustness of the proposed SP-DPC method to additive uncertainties and its capability to deal with a range of control tasks such as stabilization of unstable systems, stochastic reference tracking, and stochastic parametric obstacle avoidance with nonlinear constraints.
 
% \subsection{Mathematical Background}

% robust MPC
% stability theorems + assumptions
% all that is known already in the literature
% maybe mention connections and differences with RL and MPC?

% https://spinningup.openai.com/en/latest/spinningup/rl_intro.html#reward-and-return
% https://spinningup.openai.com/en/latest/spinningup/rl_intro3.html

% differences with RL:
% in RL they suppose that both the environment transitions and the policy are stochastic. 
% in DPC both are deterministic
% in RL they they need to learn their value functions
% in DPC we know them explicitly
% in RL they use MDPs
% in DPC we use MPC/ constrained optimal control problem formalism
% in RL environment is only for generating trajectories and is not differentiable
% in DPC the environment is differentiable

\section{Method}

% \subsection{Differentiable Robust Optimal Control Problem}

\subsection{Problem Statement}

Let us consider following uncertain discrete-time linear dynamical system:
\begin{equation}
\label{eq:truth:model:uncertain}
    {\bf x}_{k+1} = {\bf A}  {\bf x}_k + {\bf B } {\bf u}_k + \boldsymbol \omega_k,
\end{equation}
where, 
${\bf x}_k \in \mathbb{X} \subseteq \mathbb{R}^{n_{x}}$
are state variables, and ${\bf u}_k \in \mathbb{U} \subseteq \mathbb{R}^{n_{u}}$ are control inputs. The state matrices $\bf A$, and $\bf B$ are assumed to be known, however, the dynamical system is corrupted by $\boldsymbol \omega_k \in \Omega \subseteq \mathbb{R}^{n_{\omega}}$ which is time-varying additive unmeasured uncertainty.
We consider that states and inputs are subject to joint parametric nonlinear chance constraints in their standard form:
\begin{subequations}
\label{eq:chance_con}
    \begin{align}
   \textbf{Pr}( h({\bf x}_k, {\bf p}_k) \le {\bf 0}) 
   \ge  \beta,
 \\
   \textbf{Pr}(  g({\bf u}_k,   {\bf p}_k) \le {\bf 0}) \ge \beta,
 \end{align}
 \end{subequations}
where  $ {\bf p}_k \in \Xi \subset \mathbb{R}^{n_{p}}$ represents vector of parameters, and 
$\beta \in (0,1]$ is the user-specified probability requirement, \add{ with $\beta = 1$ representing hard constraint where the state and parameter dependent constraints have to satisfy over all times.} However, with relaxed probabilistic consideration, i.e., with $\beta < 1$ we can allow for constraint violation with probability $1-\beta$, thereby trading-off performance with the robustness. 
Unfortunatelly, joint chance constraints~\eqref{eq:chance_con} are
in general non-convex and intractable.
One of the approaches for dealing with joint chance constraints in constrained optimization is to use tractable
deterministic surrogates via sampling~\citep{BAVDEKAR2016270}:
\begin{subequations}
\label{eq:det_chance_con}
    \begin{align}
    h({\bf x}_k^j,   {\bf p}_k) & \le {\bf 0}, \ j \in \mathbb{N}_1^{s}
 \\
   g({\bf u}_k^j,   {\bf p}_k) & \le {\bf 0}, \ j \in \mathbb{N}_1^{s}
 \end{align}
 \end{subequations}
where $j$ represents the index of the deterministic realization of the chance constraints with uncertainties sampled from the distribution  $\boldsymbol\omega_k^j \thicksim P_{\boldsymbol \omega}$ with  $s$ number of samples.

For control, we consider an arbitrary differentiable performance metric given as a function of states, control actions, and problem parameters $  \ell( {\bf x}_k, {\bf u}_k, {\bf p}_k) $.  A standard example is the parametric reference tracking objective:
\begin{equation}
\label{eq:param_obj}
    \ell( {\bf x}_k, {\bf u}_k,  {\bf p}_k) = 
  || {\bf r}_{k} - {\bf x}_k ||_{Q_{r}}^2 + 
  || {\bf u}_k ||_{Q_{u}}^2 
\end{equation}
where the reference signal ${\bf r}_{k} \in \mathbb{X} \subset \mathbb{R}^{n_{x}}$ belongs to the parameter vector
${\bf p}_k$.
\add{ With  $|| \cdot||_Q^2$ is weighted squared $2$-norm  scaled with positive scalar-valued  factor $Q$.}

Given the system dynamics model~\eqref{eq:truth:model:uncertain}, 
our aim is to find  parametric predictive optimal control policy 
represented by a deep neural network
$ \pi_{\boldsymbol \theta}({\bf x}_0, \boldsymbol \xi): \mathbb{R}^{n_x + Nn_{p}} \to \mathbb{R}^{Nn_u} $~\eqref{eq:dnn} that minimizes the parametric control objective function~\eqref{eq:param_obj} over finite prediction horizon $N$, while satisfying the parametric chance constraints~\eqref{eq:chance_con}.
\add{$\boldsymbol {\xi} = [{\bf p}_0, \ldots, {\bf p}_{N-1} ]$ represents a forecast of the 
problem parameters, e.g., reference and constraints preview given over the  $N$-step horizon. 
Vector ${\bf x}_0 = {\bf x}(t)$ represents a full state feedback of the initial conditions measured at time $t$.}
We assume fully connected neural network policy given as:
\begin{subequations}
    \label{eq:dnn}
    \begin{align}
   {\bf U} = \boldsymbol{\pi}_{ \boldsymbol \theta}({\bf x}_0, \boldsymbol {\xi}) & =  \mathbf{W}_{L}  \mathbf{z}_L + \mathbf{b}_{L} \\
    \mathbf{z}_{l} &= \boldsymbol\sigma(\mathbf{W}_{l-1} \mathbf{z}_{l-1} + \mathbf{b}_{l-1})  \label{eq:dnn:layer}\\
    \mathbf{z}_0 &= [ {\bf x}_0^T, \boldsymbol {\xi}^T ]
 \end{align}
\end{subequations}
where ${\bf U} = [{\bf u}_0, \ldots, {\bf u}_{N-1} ]$
is an open-loop control sequence.
In the policy architecture, 
 $\mathbf{z}_i$ represent hidden states, $\mathbf{W}_i$, and $\mathbf{b}_i$ being weights and biases of the $i$-th layer, respectively, compactly represented as policy parameters $\boldsymbol \theta$ to be optimized.
The nonlinearity $\boldsymbol\sigma: \mathbb{R}^{n_{z_l}} \rightarrow \mathbb{R}^{n_{z_l}}$ is given by element-wise execution of a differentiable activation function $\sigma: \mathbb{R} \rightarrow \mathbb{R}$.

We consider the following stochastic parametric optimal control problem:
\begin{subequations}
\add{
\label{eq:SPOCP}
    \begin{align}
\min_{\boldsymbol \theta} \ & \mathbb{E} \ \Big( \sum_{k=0}^{N-1}\ell( {\bf x}_k, {\bf u}_k, {\bf p}_k) + || {\bf x}_N ||_{Q_f}^2 \Big),  & 
 \\  
  \text{s.t.} \ &   {\bf x}_{k+1} = {\bf A} {\bf x}_k + {\bf B } {\bf u}_k + \boldsymbol \omega_k, \  k \in \mathbb{N}_{0}^{N-1},     & \\
 \  &  {\bf U} = \boldsymbol{\pi}_{ \boldsymbol \theta}({\bf x}_0, \boldsymbol {\xi}), \\
  & \textbf{Pr}( h({\bf x}_k, {\bf p}_k) \le {\bf 0}) 
   \ge  \beta,
 \\
 &  \textbf{Pr}(  g({\bf u}_k,   {\bf p}_k) \le {\bf 0}) \ge \beta, \\
   \ &  {\bf x}_0 \thicksim P_{{\bf x}_0},  \\ 
  \ &\boldsymbol \xi = [{\bf p}_0, \ldots, {\bf p}_{N-1} ] \thicksim P_{\boldsymbol \xi},  \\ 
    \ &\boldsymbol \omega_k \thicksim P_{\boldsymbol \omega}, \  k \in \mathbb{N}_{0}^{N-1}.
\end{align}
}
\end{subequations}
\add{Our objective is to optimize  the explicit neural control policy~\eqref{eq:dnn} by solving the parametric problem~\eqref{eq:SPOCP}  over the given distributions of initial conditions $\thicksim P_{{\bf x}_0}$, problem parameters $P_{\boldsymbol \xi}$, and uncertainties $\thicksim P_{\boldsymbol \omega}$.
Please note that in contrast with stochastic MPC schemes we do not solve the problem~\eqref{eq:SPOCP} online, instead we compute the explicit solution offline.
Before we present our policy optimization algorithm, we consider the following assumptions.}

\textit{Assumption 1:}
The nominal model~\eqref{eq:truth:model:uncertain} is controllable.

\textit{Assumption 2:}
The parametric control performance metric $  \ell( {\bf x}_k, {\bf u}_k,  {\bf p}_k) $
and constraints $  h({\bf x}_k,   {\bf p}_k)$, and $  g({\bf u}_k,   {\bf p}_k)$, respectively, are at least once  differentiable functions.

\textit{Assumption 3:} The input disturbances are bounded, i.e., $\omega_k \in \Omega : \{ \omega_k : ||\omega_k||_\infty \leq \nu \le \infty \}, \ \forall k \in \mathbb{N}_{0}^{N-1}$.

\textit{Assumption 4: \citep{hertneck2018learning}} There exist a local Lyapunov function $V({\bf x}_k) = || {\bf x}_k ||_{Q_f}^2$ with the terminal set $\mathcal{X}_T = \{ x_k : V({\bf x}_k) \leq \kappa \}$ with control law ${\bf K}{\bf x}_k$ such that $\forall {\bf x}_k \in \mathcal{X}_T$, ${\bf A} {\bf x}_k + {\bf B } {\bf K}{\bf x}_k + \boldsymbol \omega_k \in \mathcal{X}_T$, $\omega_k \in \Omega$, and the decrease in the Lyapunov function is bounded by  $ -\ell( {\bf x}_k, {\bf u}_k,  {\bf p}_k) $ in the terminal set $\mathcal{X}_T$.

Assumption $4$ helps us to guarantee that once the chance constraints are satisfied during the transient phase of the closed-loop, and when the trajectories reach the terminal set $\mathcal{X}_T$ under the bounded disturbances, it will be robustly positive invariant. 

Table~\ref{tab:variables} summarizes the notation used in this paper.
\begin{table}[htb!]
\begin{center}
\caption{Overview of the notation used.}\label{tab:variables}
\begin{tabular}{cccc}
notation & meaning & sampled from & belongs to  \\ \hline
 ${\bf U}^{i,j}$ & control trajectories & - & $\mathbb{U}$
 % $\pi_{\boldsymbol \theta}({\bf x}_0^{i,j}, \boldsymbol \xi^i)$ & $\mathbb{U}$
 \\
${\bf x}_0^{i,j}$ & initial system states & $P_{{\bf x}_0}$ & $\mathbb{X}$ \\
${\bf p}_k^i$ & problem parameters & $P_{\boldsymbol \xi}$ & $\Xi$\\
% ${\bf V}^i$ & param. uncertainties & $P_{{\bf V}}$ & $\Psi$ \\
$\boldsymbol \omega^j_k$ & additive uncertainties & $P_{\boldsymbol \omega}$ & $\Omega$\\ 
$i$ & parametric scenario & - & $\mathbb{N}_{1}^{m}$ \\
$j$ & uncertainty scenario & - & $\mathbb{N}_{1}^{s}$ \\
$k$ & time index & - & $\mathbb{N}_{0}^{N}$ \\
\hline
\end{tabular}
\end{center}
\end{table}

\subsection{Stochastic Parametric Differentiable Predictive Control}

In this section, we present stochastic parametric differentiable predictive control (SP-DPC) method  
that is cast as the following parametric stochastic optimal control problem in the Lagrangian form:
\begin{subequations}
\label{eq:DPC}
    \begin{align}
 \min_{\boldsymbol \theta} \ &  \ J({\bf x}_0, \boldsymbol \xi,  \boldsymbol\Omega) & 
 \label{eq:DPC:objective} \\ 
  \text{s.t.} \ &   {\bf x}_{k+1}^{i,j} = {\bf A} {\bf x}_k^{i,j} + {\bf B } {\bf u}_k^{i,j} + \boldsymbol \omega_k^j, \  k \in \mathbb{N}_{0}^{N-1}   \label{eq:dpc:x}  & \\
   \  & {\bf U}^{i,j} = \boldsymbol{\pi}_{ \boldsymbol \theta}({\bf x}^{i,j}_0, \boldsymbol {\xi}^i) \label{eq:dpc:pi} \\
   \ &  {\bf x}_0^{i,j} \thicksim P_{{\bf x}_0},  \label{eq:dpc:x0} \\ 
  \ &\boldsymbol \xi^i  = [{\bf p}_0^i, \ldots, {\bf p}_{N-1}^i ] \thicksim P_{\boldsymbol \xi},  \label{eq:dpc:xi} \\ 
    \ & \boldsymbol \Omega^j = [\boldsymbol \omega_0^j,
    \ldots, \boldsymbol \omega_{N-1}^j] \thicksim P_{\boldsymbol \omega} \label{eq:dpc:omega} \\
%   \ &  {\bf V}^j \thicksim P_{{\bf V}} \label{eq:dpc:omega} \\
\  & k \in \mathbb{N}_{0}^{N-1}, \ i \in \mathbb{N}_{1}^{m}, \ j \in \mathbb{N}_{1}^{s} 
\end{align}
\end{subequations}
The stochastic evolution of the state variables ${\bf x}_k^{i,j}$ is obtained via the rollouts of the system model~\eqref{eq:dpc:x}
  with initial conditions sampled from the distribution $P_{x_0}$. Vector ${\bf p}_k^i$ represents optimal control problem parameters sampled from the distribution $P_{\boldsymbol \xi}$.
  Samples of the initial conditions~\eqref{eq:dpc:x0} and problem parameters~\eqref{eq:dpc:xi} define $m$ number of known unique environment scenarios.
On the other hand, $\boldsymbol \omega_k^j$ represents unmeasured additive  uncertainties independently sampled from distrubution $P_{\boldsymbol \omega}$, and define $s$ number of unique disturbance scenarios with each scenario leads to one uncertain episode. 

Overall, the problem~\eqref{eq:DPC} has $ms$ unique scenarios indexed by a tuple $(i,j)$  
that are parametrizing by expected value of the loss function~\eqref{eq:DPC:objective} through their effect on the system dynamics rollouts over $N$ steps via the model equation~\eqref{eq:dpc:x}.
Hence, the parametric  loss function is defined over sampled distributions of problem's initial conditions, measured time-varying parameters, and unmeasured additive and parametric uncertainties as follows:
\begin{equation}
\add{
    \begin{split}
    \label{eq:policy_loss}
    J({\bf x}_0, \boldsymbol \xi, \boldsymbol\Omega) =   \frac{1}{msN} \sum_{i=1}^{m} \sum_{j=1}^{s} \Big( \sum_{k=0}^{N-1}  \big( \ell( {\bf x}_k^{i,j}, {\bf u}_k^{i,j}, {\bf p}_k^i )  + \\ 
 p_x(h({\bf x}_k^{i,j}, {\bf p}_k^i))  +   p_u(g({\bf u}_k^{i,j}, {\bf p}_k^i))  \big) + || {\bf x}^{i,j}_N ||_{Q_f}^2 \Big)
    \end{split}
    }
\end{equation}
where  $ \ell( {\bf x}_k, {\bf u}_k, {\bf p}_k): \mathbb{R}^{n_x + n_u + n_p} \to \mathbb{R}  $ defines the control objective, while $p_x(h({\bf x}_k, {\bf p}_k)): \mathbb{R}^{n_h + n_{p}} \to \mathbb{R}   $, and 
$ p_u(g({\bf u}_k, {\bf p}_k)): \mathbb{R}^{n_g + n_{p}} \to \mathbb{R} $ define penalties of parametric
 state and input constraints, defined as:
\begin{subequations}
\label{eq:ReLU_ineq}
    \begin{align}
    p_x(h({\bf x}_k,  {\bf p}_k) & =
  ||
    \texttt{ReLU}(h({\bf x}_k,  {\bf p}_k)||_{Q_h}^2 \\
p_u(g({\bf u}_k,  {\bf p}_k )  & =   || \texttt{ReLU}(g({\bf u}_k,  {\bf p}_k)||_{Q_g}^2
 \end{align}
 \end{subequations}
 \add{where  $|| \cdot||_Q^2$ represents squared $2$-norm weighted by a positive scalar factor $Q$, and $\texttt{ReLU}$ stands for rectifier linear unit  function given as $\texttt{ReLU}(x) = \text{max}(0,x)$. }

\subsection{Stochastic Parametric DPC Policy Optimization}

The advantage of the model-based approach of DPC is that we can directly compute the policy gradient using automatic differentiation.
For a simpler exposition of the policy gradient, we start by defining the following proxy variables:
\begin{subequations}
\begin{align}
    L & = \frac{1}{msN} \sum_{i=1}^{m} \sum_{j=1}^{s} \sum_{k=0}^{N-1}\ell( {\bf x}_k^{i,j}, {\bf u}_k^{i,j}, {\bf p}_k^i) \\
    L_x & = \frac{1}{msN} \sum_{i=1}^{m} \sum_{j=1}^{s} \sum_{k=0}^{N-1} p_x(h({\bf x}_k^{i,j}, {\bf p}_k^i)) \\
    L_u & = \frac{1}{msN} \sum_{i=1}^{m} \sum_{j=1}^{s} \sum_{k=0}^{N-1} p_u(g({\bf u}_k^{i,j}, {\bf p}_k^i))
\end{align}
\end{subequations}
\add{where $L$, $L_x$, and $L_u$} represent scalar values of the  control objective, state, and input constraints, respectively,  evaluated on the $N$-step ahead rollouts of the closed-loop system dynamics over the batches of sampled problem parameters ($i$-th index) and uncertainties ($j$-th index). This parametrization now allows us to express the policy gradient by using the chain rule as follows:
\begin{equation}
\label{eq:grad}
\add{
\begin{split}
    \nabla_{\boldsymbol \theta} J =  
    \frac{ \partial L}{\partial\boldsymbol \theta} +  \frac{ \partial L_x}{\partial \boldsymbol \theta} +
    \frac{ \partial L_u }{\partial \boldsymbol \theta} = \\
    \frac{ \partial L}{\partial {\bf x}}   \frac{ \partial {\bf x}}{\partial {\bf u}} \frac{ \partial {\bf u}}{\partial \boldsymbol \theta} +   \frac{ \partial L}{\partial {\bf u}}  \frac{ \partial {\bf u}}{\partial \boldsymbol \theta} +  
    \frac{ \partial L_x}{\partial {\bf x}}  \frac{ \partial {\bf x}}{\partial {\bf u}} \frac{ \partial {\bf u}}{\partial \boldsymbol \theta} +   \frac{ \partial L_u}{\partial {\bf u}} \frac{ \partial {\bf u}}{\partial \boldsymbol \theta}
\end{split}
}
\end{equation}
\add{Here $\frac{ \partial {\bf u}}{\partial \boldsymbol \theta}$ are
  partial derivatives of the actions with respect to the  policy parameters that can be computed via  backpropagation through the neural architecture.}

Having fully parametrized policy gradient~\eqref{eq:grad} now allows us to train the neural control policies offline to obtain the  approximate solution of the stochastic parametric optimal control problems~\eqref{eq:DPC}  using gradient-based optimization.
Specifically, we  propose a policy optimization algorithm~\eqref{algo:DPC_optim} that is based on joint sampling of the problem parameters and initial conditions from distributions $P_{\boldsymbol \xi}$, and $P_{{\bf x}_0}$, creating $m$ parametric scenarios, and independent sampling of the uncertainties from the distribution $P_{\boldsymbol \omega}$, creating $s$ uncertainty scenarios.
Then we construct a differentiable computational graph of the problem~\eqref{eq:DPC}  with known matrices of the nominal system model~\eqref{eq:truth:model:uncertain}, randomly initialized neural control policy~\eqref{eq:dnn}, and parametric  loss function~\eqref{eq:policy_loss}.
This model-based approach now allows us to directly compute the policy gradients~\eqref{eq:grad} by backpropagating the loss function values~\eqref{eq:policy_loss}  through the unrolled closed-loop system dynamics model over the $N$ step prediction horizon window. 
The computed policy gradients are then used to train the weights of the control policy $\boldsymbol \theta$ via gradient-based optimizer $\mathbb{O}$, e.g., stochastic gradient descent and its popular variants.
\begin{algorithm}[htb!]
  \caption{Stochastic parametric differentiable predictive control (SP-DPC) policy optimization.}\label{algo:DPC_optim}
  \begin{algorithmic}[1]
  \State \textbf{input} training datasets with $m$ samples of  initial conditions ${\bf x}_0$ and problem parameters  $\boldsymbol \xi$ sampled from the distributions $P_{{\bf x}_0}$, and $P_{\boldsymbol \xi}$, respectively
\State \textbf{input} training datasets of $s$ realizations of uncertainties $\boldsymbol \Omega$ sampled from the distribution $P_{\boldsymbol \omega}$
  \State \textbf{input} nominal system model $({\bf A} ,{\bf B })$
 \State \textbf{input}  neural feedback policy architecture $\pi_{\boldsymbol \theta}({\bf x}_0, \boldsymbol \xi)$
  \State \textbf{input} stochastic parametric DPC loss function  $J$~\eqref{eq:policy_loss}
   \State \textbf{input} gradient-based optimizer $\mathbb{O}$
 \State \textbf{differentiate}  parametric  DPC loss $J$~\eqref{eq:policy_loss} over the 
 sampled distributions of the initial conditions, problem parameters, and uncertainties
 to obtain the policy gradient $ \nabla_{{\bf W}} J $~\eqref{eq:grad}
 \State \textbf{learn} policy $\pi_{\boldsymbol \theta}$  via optimizer  $\mathbb{O}$ using gradient $ \nabla_{\boldsymbol \theta} J $
\State \textbf{return} optimized parameters $\boldsymbol \theta$ of the policy $\pi_{ \boldsymbol \theta}$
  \end{algorithmic}
\end{algorithm}

% \textit{Remark 1}:
The Algorithm~\ref{algo:DPC_optim} can be implemented in any programming language   
supporting reverse mode automatic differentiation on user-defined computational graphs, e.g., Julia, Swift, or Jax. In our case, we implemented the proposed algorithm using Pytorch~\citep{paszke2019pytorch} in Python.

% \textit{Remark 1}:
% Thanks to the known system dynamics model, and known parametric forms of the control objective and chance constraints in the proposed policy optimization Algorithm~\ref{algo:DPC_optim}, there is no need for value approximation based on explicit reward signal feedback as in the case of actor critic reinforcement learning (RL) algorithms. Instead, the parametric Lagrangian loss~\eqref{eq:policy_loss} represents the value function that can be directly sampled without approximation error.

% \subsection{Learning Parametric Lyapunov Functions}
% parametrize: 1, references 2, uncertainties
% use simple change of basis to parametrize the attractor of the lyapunov function to reflect the reference signal to minimize tracking error e = r - x
% % https://en.wikipedia.org/wiki/Change_of_basis

% TODO: we can explore to CLF or storage functions later
% % https://en.wikipedia.org/wiki/Lyapunov_function
% % https://en.wikipedia.org/wiki/Control-Lyapunov_function
% % https://en.wikipedia.org/wiki/Dissipative_system

% TODO: explore lit on parametric lyapunov functions
% https://www.sciencedirect.com/science/article/abs/pii/S0167691196000758
% https://citeseerx.ist.psu.edu/viewdoc/download?doi=10.1.1.489.8837&rep=rep1&type=pdf
% https://ieeexplore.ieee.org/document/486646
% https://dcsl.gatech.edu/papers/cdc03a.pdf

% TODO: for system ID - lyapunov functions with multiple equilibiria
% https://www.sciencedirect.com/science/article/pii/S1877050917309377
% https://www.researchgate.net/publication/275250678_Computation_of_Lyapunov_functions_for_systems_with_multiple_local_attractors

\section{Probabilistic Guarantees}

In a learning-based MPC setting, recent works such as \citep{hertneck2018learning, rosolia2018stochastic, Karg2021} discuss a few probabilistic considerations. In this paper, we bring in a novel stochastic sampling-based design for differentiable predictive control architecture along with chance constraints for closed-loop state evolution and provide appropriate probabilistic guarantees motivated from \citep{hertneck2018learning}. We use the differentiable predictive control policy optimization Algorithm~\ref{algo:DPC_optim}  to solve the stochastic optimization problem~\eqref{eq:DPC}. The algorithm uses a forward simulation of the trajectories, which follows the uncertain dynamical behavior. The system is perturbed by the bounded disturbance input $\omega_k^j$. Therefore, for a given initial condition, the uncertainty sources belong to
$ \boldsymbol \omega_k^j \in \Phi,$ where $\Phi$ represents a bounded set.
Moreover, during the batch-wise training, we have considered different trajectories belonging to
different parametric scenarios with
varying initial conditions ${\bf x}_0^i \in \mathbb{X} \subseteq \mathbb{R}^{n_{x}}$
and problem parameters $ \boldsymbol \xi^i  \in  \boldsymbol \Xi \subseteq \mathbb{R}^{Nn_{p}}$. We considered the variation of initial conditions ~\eqref{eq:dpc:x0} and different parametric scenarios~\eqref{eq:dpc:xi} contain $m$ number of known unique environment scenarios during the training, denoted with superscript $i$, and variations in different disturbance scenarios are denoted with superscript $j$ with $s$ different scenarios.

% However, theoretically it is not straightforward to consider such amount of uncertainties to provide feasibility and stability considerations for the closed-loop. Therefore, we supplement the learning problem with few specified considerations to provide probabilistic guarantees. First we add the terminal constraint on the states such that,
% \begin{align}
% \label{eq:terminal}
%     {\bf x}_{N} \in  \mathcal{X}_f.
% \end{align}
% Under uncertainties due to external disturbances, system parameters and initial conditions, the solution trajectories from \eqref{eq:dpc:x} may violate state and parameter dependent constraints. Therefore, in this paper, we adopt chance constraint  framework~\eqref{eq:chance_con}.

Before discussing the feasibility and stability theorem, let us denote the set of state trajectories over the stochastic closed-loop system rollouts  ${\bf X}^{i,j}$, 
with control action sequences generated using the learned explicit SP-DPC policy $[{\bf u}_0^{i,j}, \ldots, {\bf u}_{N-1}^{i,j}] = \pi_{\boldsymbol \theta}({\bf x}_0^{i,j}, \boldsymbol \xi^i)$,
for given samples of parametric $\ i \in \mathbb{N}_{1}^{m}$ and disturbance $j \in \mathbb{N}_{1}^{s} $ scenarios,  i.e.,
\begin{align}
    {\bf X}^{i,j}: \begin{cases} &\boldsymbol \{ {\bf x}_k^{i,j}\}, \ \forall k \in \mathbb{N}_{0}^{N-1},   \nonumber\\
    &{\bf x}_{k+1}^{i,j} = {\bf A} {\bf x}_k^{i,j} + {\bf B } {\bf u}_k^{i,j} + \boldsymbol \omega_k^j. 
    \end{cases}
\end{align}
We also compactly denote the constraints satisfaction metric over the sampled trajectories,
\begin{align}
    {\bf P}^{i,j}= \mbox{True} : \begin{cases}
   h({\bf x}_k^{i,j}, {\bf p}_k^i) \le {\bf 0}, \ \forall k \in \mathbb{N}_{0}^{N-1},
 \\
  g({\bf u}_k^{i,j},  {\bf p}_k^i) \le {\bf 0}, \ \forall k \in \mathbb{N}_{0}^{N-1}, \\
  {\bf x}_N^{i,j} \in \mathcal{X}_T
    \end{cases}
\end{align}
Now we define the indicator function,
\begin{align}
    \mathcal{I}({\bf X}^{i,j}) := \begin{cases}
    \label{eq:Indicator}
    1 \;\; \text{if} \;\; {\bf P}^{i,j}=\mbox{True}.\\
    0, \;\; \text{otherwise},
    \end{cases}
\end{align}
which signifies whether the learned control law $\pi_{\boldsymbol \theta}({\bf x}_0^{i,j}, \boldsymbol \xi^i)$ satisfies the constraints along a sample trajectory until the terminal set $\mathcal{X}_T$ is reached.

\textbf{Theorem 1:} 
\label{thm:theorem1}
Consider the sampling-based approximation of the stochastic parametric constrained optimal control problem~\eqref{eq:DPC} along with assumptions $1-4$, and the SP-DPC policy optimization Algorithm~\ref{algo:DPC_optim}. Choose chance constraint violation probability $1-\beta$, and level of confidence parameter $\delta$.
Then if the empirically computed risk $\tilde{\mu}$ on the indicator function \eqref{eq:Indicator} with sufficiently large number of sample trajectories $r=ms$ satisfy $\beta \le
    \tilde{\mu} - \sqrt{-\frac{\ln{\frac{\delta}{2}}}{2r}}$, then  the learned SP-DPC policy $\pi_{\boldsymbol \theta}({\bf x}_0, \boldsymbol \xi)$ will guarantee satisfaction of the chance constraints~\eqref{eq:chance_con} and closed-loop stability in probabilistic sense.  \qed

\textit{Proof:} We consider $m$ different scenarios encompassing sampled initial conditions and parameters, and $s$ different disturbance scenarios, thereby generating $r = ms$ sample trajectories. Moreover, the initial conditions, parameters and disturbance scenarios are each sampled in an iid fashion, thereby making ${\bf X}^{i,j},\mathcal{I}({\bf X}^{i,j})$ iid with the learned control policy $\pi_{\boldsymbol \theta}({\bf x}_0^{i,j}, \boldsymbol \xi^i)$. The empirical risk over trajectories is defined as,
\begin{align}
    \Tilde{\mu} = \frac{1}{ms}\sum_{i=1}^m\sum_{j=1}^s \mathcal{I}({\bf X}^{i,j}).
\end{align}

The constraint satisfaction and stability are guaranteed if for ${\bf x}_0^{i,j} \thicksim P_{{\bf x}_0}, 
  \ \boldsymbol \xi^i  \thicksim P_{\boldsymbol \xi}, 
    \ \boldsymbol \omega_k^j \thicksim P_{\boldsymbol \omega}$, we have, $\mathcal{I}({\bf X}^{i,j}) = 1$, i.e., deterministic samples of chance constraints are satisfied, $h({\bf x}_k^{i,j},  {\bf p}_k^i) \le {\bf 0}$,
  $  g({\bf x}_k^{i,j},  {\bf p}_k^i) \le {\bf 0}$, ${\bf x}_N^{i,j} \in \mathcal{X}_T$, $\forall i \in \mathbb{N}_1^m$, and $\forall j \in \mathbb{N}_1^s$.
   
   \add{Denoting $\mu := \textbf{Pr}(\mathcal{I}({\bf X}) = 1)$} we recall Hoeffding’s Inequality \citep{hertneck2018learning} to estimate $\mu$ from the empirical risk $\tilde{\mu}$ leading to:
%\textit{Lemma 1} [Hoeffding’s Inequality]:    
% Let $\mathcal{I}(\bf X^{i})$ be $r$ iid random variables with $0 \leq \mathcal{I}(\bf X^{i}) \leq 1$. One can have,
\begin{align}
    \textbf{Pr}(|\tilde{\mu} - \mu| \geq \alpha) \leq 2\mbox{exp}(-2r\alpha^2) \;\; \forall \alpha > 0.
\end{align}

Therefore, denoting $\delta := 2\mbox{exp}(-2r\alpha^2)$, with confidence $1 - \delta$ we will have,
\add{
\begin{align}
\label{eq:prob_guarantee}
   \textbf{Pr}(\mathcal{I}({\bf X}) = 1) = \mu \geq \tilde{\mu} - \alpha.
\end{align}}

Thus for a chosen confidence $\delta$ and risk lower bound
\add{$\beta \leq \textbf{Pr}(\mathcal{I}({\bf X}) = 1)$}, we can evaluate the empirical risk bound:
\begin{equation}
\label{eq:mu_bound}
   \beta \le \tilde{\mu} - \alpha =
    \tilde{\mu} - \sqrt{-\frac{\ln{\frac{\delta}{2}}}{2r}}
\end{equation}
Thereby, for a fixed chosen level of confidence $\delta$ and risk lower bound $\beta$, the empirical risk $\tilde{\mu}$ and $\alpha$ can be computed for an experimental value of $r$. Thereby when \eqref{eq:mu_bound} holds for the policies trained via Algorithm~\ref{algo:DPC_optim}, then with confidence at least $1-\delta$, for at least a fraction of $\beta$ trajectories \add{${\bf X}$ we will have ${\bf P}=\mbox{True}$.} Or in other words, the chance constraints~\eqref{eq:chance_con} are satisfied with the confidence $1-\delta$.
Furthermore, along with the constraint satisfaction of the closed-loop trajectories, assumption $4$ guides us to the existence of a positive invariant terminal set in presence of bounded disturbances, thereby maintaining stability once the constraints are satisfied in probabilistic sense. This concludes the proof guaranteeing  stability and constraint satisfaction of the learned policy using the SP-DPC optimization algorithm  Algorithm~\ref{algo:DPC_optim}.  \qed

% \add{add verification algorithm for safe learning}

\section{Numerical Case Studies}

In this section, we present three numerical case studies for showcasing the flexibility of the proposed SP-DPC method.
In particular, we demonstrate stochastically robust stabilization of an unstable system, reference tracking, and scalability to stochastic systems with a larger number of states and control inputs,
as well as stochastic parametric obstacle avoidance with nonlinear constraints.
All examples are implemented in the NeuroMANCER~\citep{Neuromancer2021}, which is a Pytorch-based~\citep{paszke2019pytorch} toolbox for solving constrained parametric optimization problems using sampling-based algorithms such as the proposed Algorithm~\ref{algo:DPC_optim}.
All examples below have been trained using the stochastic gradient descent optimizer AdamW~\citep{loshchilov2017decoupled}.
\add{
All experiments were performed on a 64-bit machine with 2.60 GHz Intel(R) i7-8850H CPU and 16 GB RAM.}

\subsection{Stabilizing Unstable Constrained Stochastic System}
 \label{sec:ex_1_stabilize}

In this example, we show the ability of the SP-DPC policy optimization Algorithm~\ref{algo:DPC_optim} to learn offline the stabilizing neural feedback policy for an unstable double integrator:
 \begin{equation}
 \label{eq:double_int_un}
{\bf x}_{k+1} = \begin{bmatrix} 1.2 & 1.0 \\ 0.0 & 1.0\end{bmatrix}  {\bf x}_k + \begin{bmatrix} 1.0  \\ 0.5  \end{bmatrix}   {\bf u}_k 
+ \boldsymbol \omega_k
 \end{equation}
 with normally distributed additive uncertainties $\boldsymbol \omega_k \thicksim \mathcal{N}(0, 0.01)$.
 For stabilizing the system~\eqref{eq:double_int_un} we consider the following quadratic control performance metric:
 \begin{equation}
 \add{
 \label{eq:empc_qp}
    \ell( {\bf x}_k, {\bf u}_k, {\bf p}_k) =  
 || {\bf x}_k||_{Q_x}^2  + || {\bf u}_k ||_{Q_u}^2 
 }
 \end{equation}
 with the prediction horizon $N=2$.
We also consider the following static state and input constraints:
 \begin{subequations}
\label{eq:double_int_constr}
    \begin{align}
  h({\bf x}_k, {\bf p}_k) & : \ -{\bf x}_k -{\bf 10}  \le  {\bf 0} , \ {\bf x}_k -  {\bf 10} \le  {\bf 0} 
 \\
  g({\bf u}_k, {\bf p}_k) & : \ -{\bf u}_k -{\bf 1}  \le  {\bf 0} , \ {\bf u}_k -  {\bf 1} \le  {\bf 0}  
 \end{align}
 \end{subequations}
Additionally, we consider the terminal constraint:
\begin{equation}
\label{eq:terminal_con}
    {\bf x}_N \in \mathcal{X}_f: -{\bf 0.1} \le {\bf x}_N \le {\bf 0.1}.
\end{equation}
We use known system dynamics matrices~\eqref{eq:double_int_un}, the control metric~\eqref{eq:empc_qp}, and constraints~\eqref{eq:double_int_constr}, and~\eqref{eq:terminal_con}, the loss function $J$~\eqref{eq:policy_loss}
using the constraints penalties~\eqref{eq:ReLU_ineq}.
For relative weights of the loss function terms, we use $Q_x = 5.0$, 
$Q_u = 0.2$, $Q_h = 10.0$,  $Q_g = 100.0$, $Q_f = 1.0$, where $Q_f$ refers to the terminal penalty weight.
This allows us to use a fully parametrized loss function $J$
in Algorithm~\eqref{algo:DPC_optim} to train the full-state feedback neural policy ${\bf u}_k = \pi_{\boldsymbol \theta}({\bf x}_k): \mathbb{R}^{2} \to \mathbb{R}^{1}$  with $4$ layers, each with $20$-hidden unites, and \texttt{ReLU} activations functions. 
 For generating the synthetic training dataset we use $m=3333$ samples of the uniformly distributed initial conditions  from the interval ${\bf x}_0^i \in [-{\bf 10}, {\bf 10}]$.
Then for each initial condition  ${\bf x}_0^i$ we sample $s=10$ realizations of the uncertainties $\boldsymbol \omega_k^j$ and form the dataset with total number of \add{$sm = 33330$} iid samples.
 Thus in this example, the problem parameters represent the initial conditions $\boldsymbol \xi = [{\bf x}_0^T]$.
Then for the given number of samples $r=ms$ and a chosen confidence $\delta=0.99$ we evaluate the empirical risk~\eqref{eq:mu_bound}. Thus we can say with $99.0 \%$  confidence that we  satisfy the stability and  constraints  with probability at least  $\Tilde{\mu} -\alpha  = 0.9951 $.

\begin{figure}[htb!]
    \centering
    % \hspace{-0.3cm}
    \includegraphics[width=.40\textwidth]{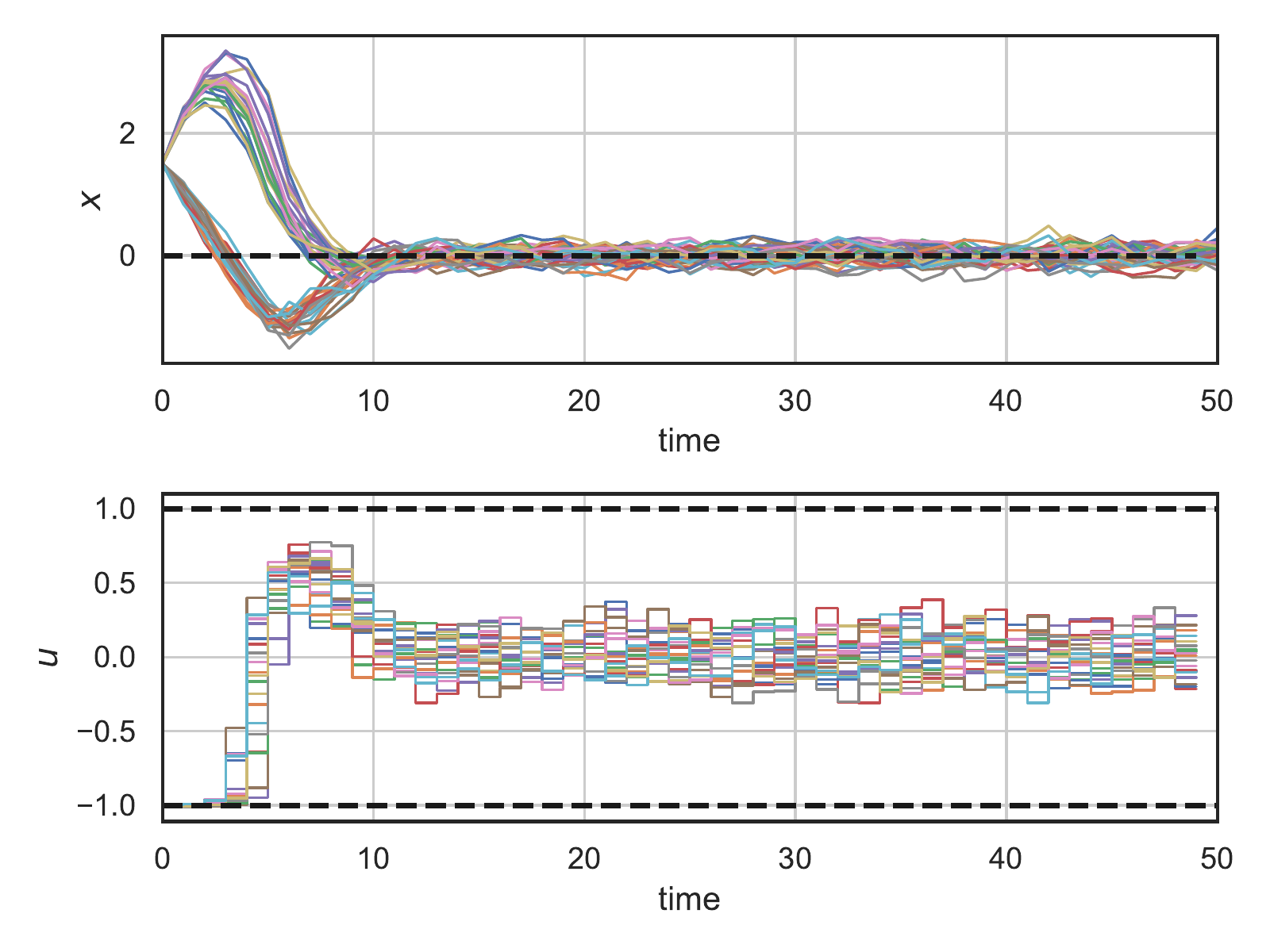}
    \caption{Closed-loop trajectories of the stochastic double integrator system~\eqref{eq:double_int_un} controlled by stabilizing neural feedback policy trained using  Algorithm~\ref{algo:DPC_optim} with SP-DPC problem formulation~\eqref{eq:DPC}. Different colors represent   $j$-th realization of additive uncertainty scenario. }
    \label{fig:DPC_cl_ex1}
\end{figure}

The stochastic closed-loop control trajectories of the system~\eqref{eq:double_int} controlled with trained neural feedback policy and $20$ realizations of the additive uncertainties are shown in Figure~\ref{fig:DPC_cl_ex1}. In this example, we demonstrate stochastically robust control performance of the stabilizing  neural policy trained using SP-DPC policy optimization Algorithm~\ref{algo:DPC_optim}.

\subsection{Stochastic Constrained Reference Tracking }
 \label{sec:ex_2_tracking}

In this case study, we demonstrate the scalability of the proposed SP-DPC policy optimization Algorithm~\ref{algo:DPC_optim}
considering systems with a larger number of states and control actions.
In particular, we consider the
linear quadcopter model\footnote{https://osqp.org/docs/examples/mpc.html} with
 ${\bf x}_k \in \mathbb{R}^{12}$, and  ${\bf u}_k \in \mathbb{R}^4$, subject to additive uncertainties  $\boldsymbol \omega_k \in \mathbb{R}^{12} \thicksim \mathcal{N}(0, 0.02^2)$.

The objective is to track the reference with the $3$-rd state \add{${\bf y}_k = {\bf x}_{3,k}$}, while keeping the rest of the states \add{$\hat{\bf x}_k = {\bf x}_{\mathbb{N}_1^{12} \setminus 3, k}$} stable.
Hence, for training the policy via Algorithm~\ref{algo:DPC_optim}, we consider the SP-DPC problem~\eqref{eq:DPC} with quadratic control objective:
\begin{equation}
\add{
 \label{eq:reference}
   \ell({\bf x}_k, {\bf u}_k, {\bf p}_k ) =  
 || {\bf y}_k - {\bf r}_k||_{Q_r}^2  + 
  || \hat{\bf x}_k ||_{Q_x}^2
  }
 \end{equation}
with horizon $N =10$, weights $Q_r = 20$, $Q_x = 5$.
We consider the following state and input constraints:
 \begin{subequations}
\label{eq:quadcopter_constr}
    \begin{align}
  h({\bf x}_k, {\bf p}_k) & : \ -{\bf x}_k -{\bf 10}  \le  {\bf 0} , \ {\bf x}_k -  {\bf 10} \le  {\bf 0} 
 \\
  g({\bf u}_k, {\bf p}_k) & : \ -{\bf u}_k -{\bf 1}  \le  {\bf 0} , \ {\bf u}_k -  {\bf 2.5} \le  {\bf 0}  
 \end{align}
 \end{subequations}
penalized via~\eqref{eq:ReLU_ineq} with  weights $Q_{h} = 1$, and $Q_{g} = 2$, respectively.
To promote the stability, we impose the contraction constraint with penalty weight $Q_{c} = 1$:
 \begin{equation}
\label{eq:state_contract_con}
   c({\bf x}_k, {\bf p}_k) : \   || {\bf x}_{k+1} ||_p \le 0.8 || {\bf x}_{k} ||_p
\end{equation}

We trained the  neural policy~\eqref{eq:dnn} $ \pi_{\boldsymbol\theta}({\bf x}): \mathbb{R}^{12} \to \mathbb{R}^{N \times 4}$ with $2$ layers, $100$ hidden states, and \texttt{ReLU} activation functions \add{resulting in $15440$ trainable parameters}.
We  used a training dataset with $m=3333$  uniformly sampled initial conditions from the interval ${\bf x}_0^i \in [-{\bf 2}, {\bf 2}]$, with $s=10$ samples of uncertainties $\boldsymbol \omega_k \thicksim \mathcal{N}(0, 0.02^2)$ per each parametric scenario.
Then for the given number of samples $r=ms$ and a chosen confidence $\delta=0.99$ we evaluated the empirical risk bound  via~\eqref{eq:mu_bound}. Thus we can say with $99.0 \%$  confidence that we  satisfy the  stability and constraints  with probability at least  $\Tilde{\mu} -\alpha = 0.9832$.

\begin{figure}[htb!]
    \centering
    % \hspace{-0.3cm}
    \includegraphics[width=.40\textwidth]{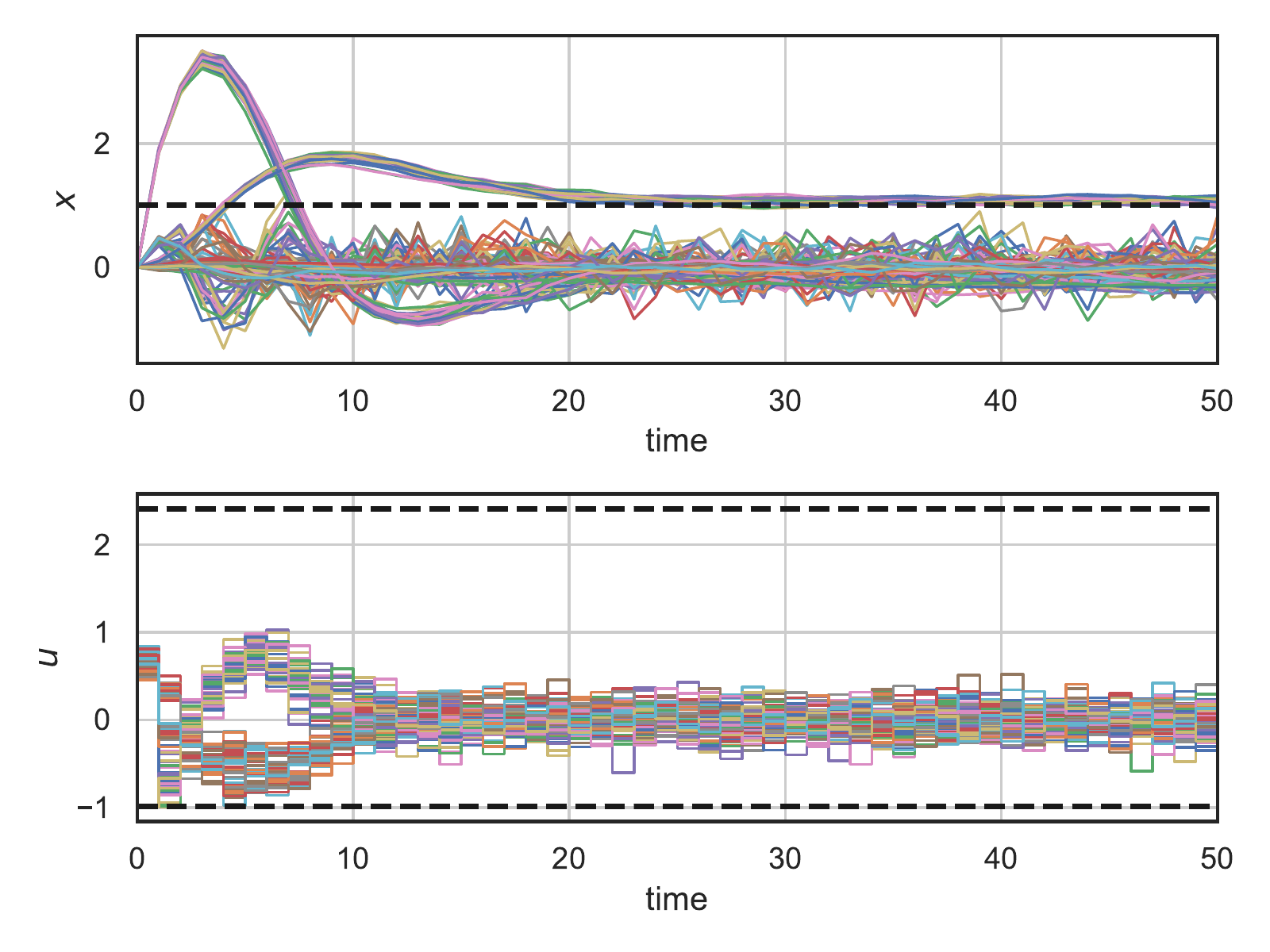}
    \caption{Closed-loop trajectories of the stochastic quadcopter model controlled by reference tracking neural feedback policy trained using  Algorithm~\ref{algo:DPC_optim} with SP-DPC problem formulation~\eqref{eq:DPC}. Different colors represent   $j$-th realization of additive uncertainty scenario. }
    \label{fig:e1:DPC_cl}
\end{figure}

Figure~\ref{fig:e1:DPC_cl} then shows the closed-loop simulations with the trained policy implemented using receding horizon control (RHC).
We demonstrate robust performance in tracking the desired reference for the stochastic system while keeping the overall system stable under perturbation.
In Table~\ref{tab:cpu_quadcopter}  we demonstrate the computational scalability of the proposed approach compared to 
implicit deterministic MPC implemented in  CVXPY~\citep{diamond2016cvxpy} and solved online via  OSQP solver~\citep{osqp2020}.
On average, we outperform the online deterministic MPC by roughly an order of magnitude in mean and maximum evaluation time.
\add{The  training time via Algorithm~\ref{algo:DPC_optim}  with $1000$ epochs took $27.71$ minutes.}
Please note that due to the larger prediction horizon and state and input dimensions, the problem being solved is far beyond the reach of classical parametric programming solvers. 
\begin{table}[htb!]
\begin{center}\caption{Comparison of online computational time of the proposed SP-DPC policy  against implicit MPC solved with OSQP.}
    \label{tab:cpu_quadcopter}
\begin{tabular}{l|ll}
{online evaluation time} &   mean  [$1e^{-3}$ s] & max  [$1e^{-3}$ s]   \\ \hline
SP-DPC &   0.272 & 1.038 \\
MPC (OSQP) &  9.196 &   82.857  \\
\hline
\end{tabular}
\end{center}
\end{table}

\subsection{Stochastic Parametric Obstacle Avoidance}
 \label{sec:ex_3_obstacle}

Here we demonstrate that the proposed SP-DPC policy optimization Algorithm~\ref{algo:DPC_optim} can be applied to 
stochastic parametric obstacle avoidance problems with nonlinear constraints.
We assume the double integrator system:
\begin{equation}
 \label{eq:double_int}
{\bf x}_{k+1} = \begin{bmatrix} 1.0 & 0.1 \\ 0.0 & 1.0\end{bmatrix}  {\bf x}_k + \begin{bmatrix} 1.0 & 0.0 \\ 0.0 & 1.0  \end{bmatrix}   {\bf u}_k 
+ \boldsymbol \omega_k
 \end{equation}
 with ${\bf x} \in \mathbb{R}^2$ and ${\bf u} \in \mathbb{R}^2$ subject to the  box constraints~\eqref{eq:double_int_constr}. 
Furthermore, let's assume an obstacle parametrized by the
 nonlinear state constraints:
\begin{equation}
 \label{eq:obstacle}
  h({\bf x}_k, {\bf p}_k) : p^2 \le b ({\bf x}_{1,k} - c)^ 2 + ({\bf x}_{2,k} - d) ^ 2
 \end{equation}
where 
${\bf x}_{i,k}$ depicts the $i$-th state, and
$p$, $b$, $c$, $d$ are parameters defining the volume, shape, and center of the  obstacle, respectively. 
For training the SP-DPC neural policy via Algorithm~\ref{algo:DPC_optim}
we use the following objective: 
\begin{equation}
\add{
 \label{eq:obstacle_loss}
 \begin{split}
J({\bf x}_0, \boldsymbol \xi,  \boldsymbol\Omega) =  \frac{1}{msN} \sum_{i=1}^{m} \sum_{j=1}^{s}
\Big( || {\bf x}_N - {\bf r}_N||_{Q_r}^2  +   \\ \sum_{k=0}^{N-2} || {\bf u}_{k+1} - {\bf u}_k ||_{Q_{du}}^2 +
  \sum_{k=0}^{N-1}  \big(
  || {\bf x}_{k+1} - {\bf x}_k ||_{Q_{dx}}^2 + 
  || {\bf u}_k ||_{Q_{u}}^2
 \big) \Big)
 \end{split}
 }
 \end{equation}
 with the prediction horizon $N=20$.
The first term penalizes the deviation of the terminal state from the target position parametrized by ${\bf r}_N$. While the second and third terms of the objective penalize the  
change in control actions and states, thus promoting solutions with smoother trajectories. The last term is an energy  minimization term on actions.
We assume the following scaling factors $Q_r = 1.0$,  $Q_{du} = 1.0$, $Q_{dx} =1.0$, $Q_u = 1.0$, and $Q_h = 100.0$, for reference, input and state smoothing, energy minimization, and constraints penalties, respectively. The uncertainty scenarios $\boldsymbol \omega_k^j $ are sampled from the normal distribution $\mathcal{N}(0, 0.01^2)$, with total number of samples $s = 100$.
The vector of sampled parameters for this problem is given as $\boldsymbol\xi^i = [{\bf x}_0^T, {\bf r}_N^T, p, b, c, d]^i$, with total number of samples $m=1000$. Thus having total number of scenarios $r=100k$
out of which $33k$ is used for training.
\add{The control policy $ \pi_{\boldsymbol \theta}(\boldsymbol\xi): \mathbb{R}^{8} \to \mathbb{R}^{N \times 2}$ is parametrized by $4$-layer $\texttt{ReLU}$
neural network with $35140$ trainable parameters.}

\begin{figure}[htb!]
\centering
% \centering
    \includegraphics[width=0.49\linewidth]{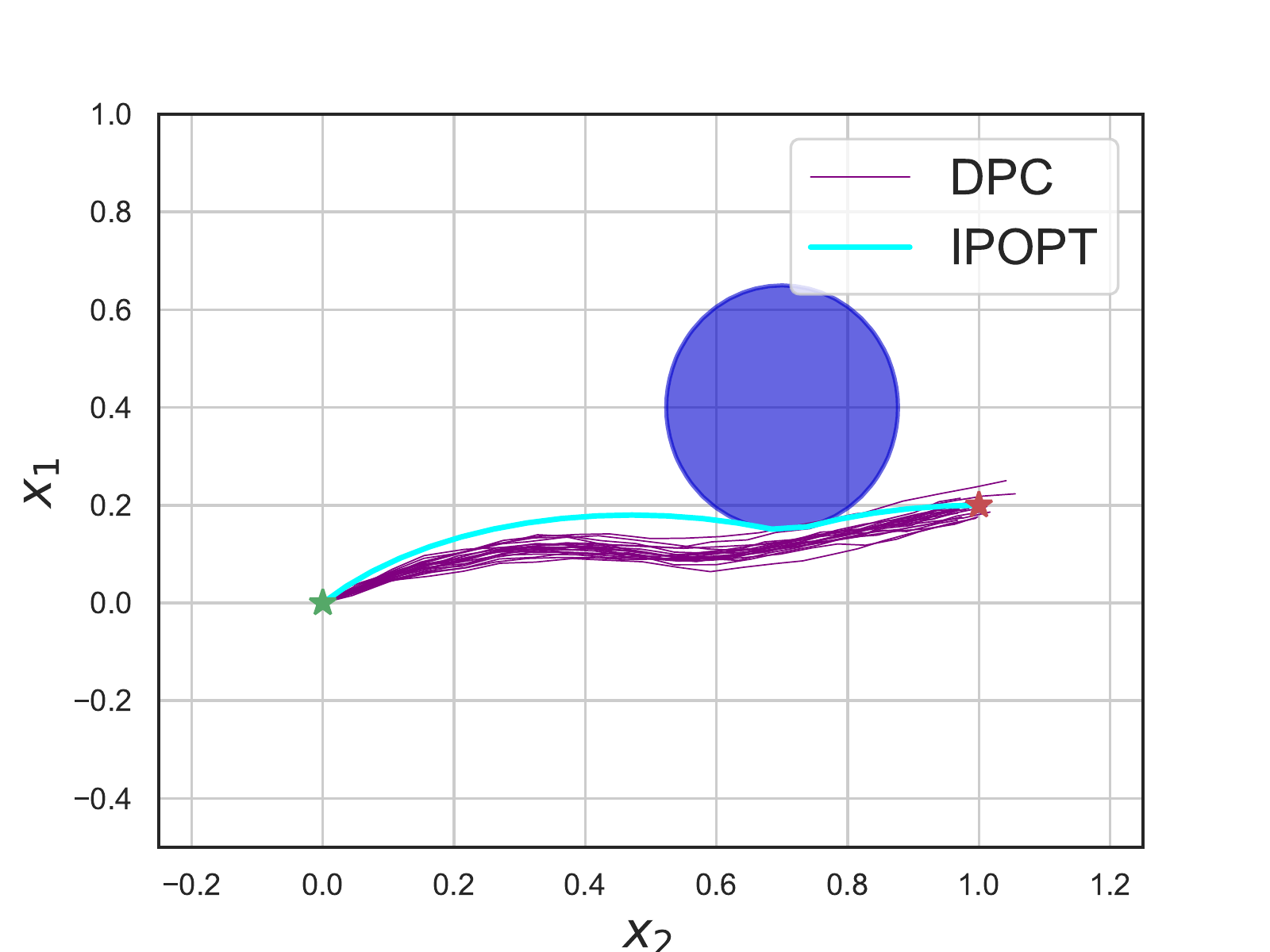}
        \includegraphics[width=0.49\linewidth]{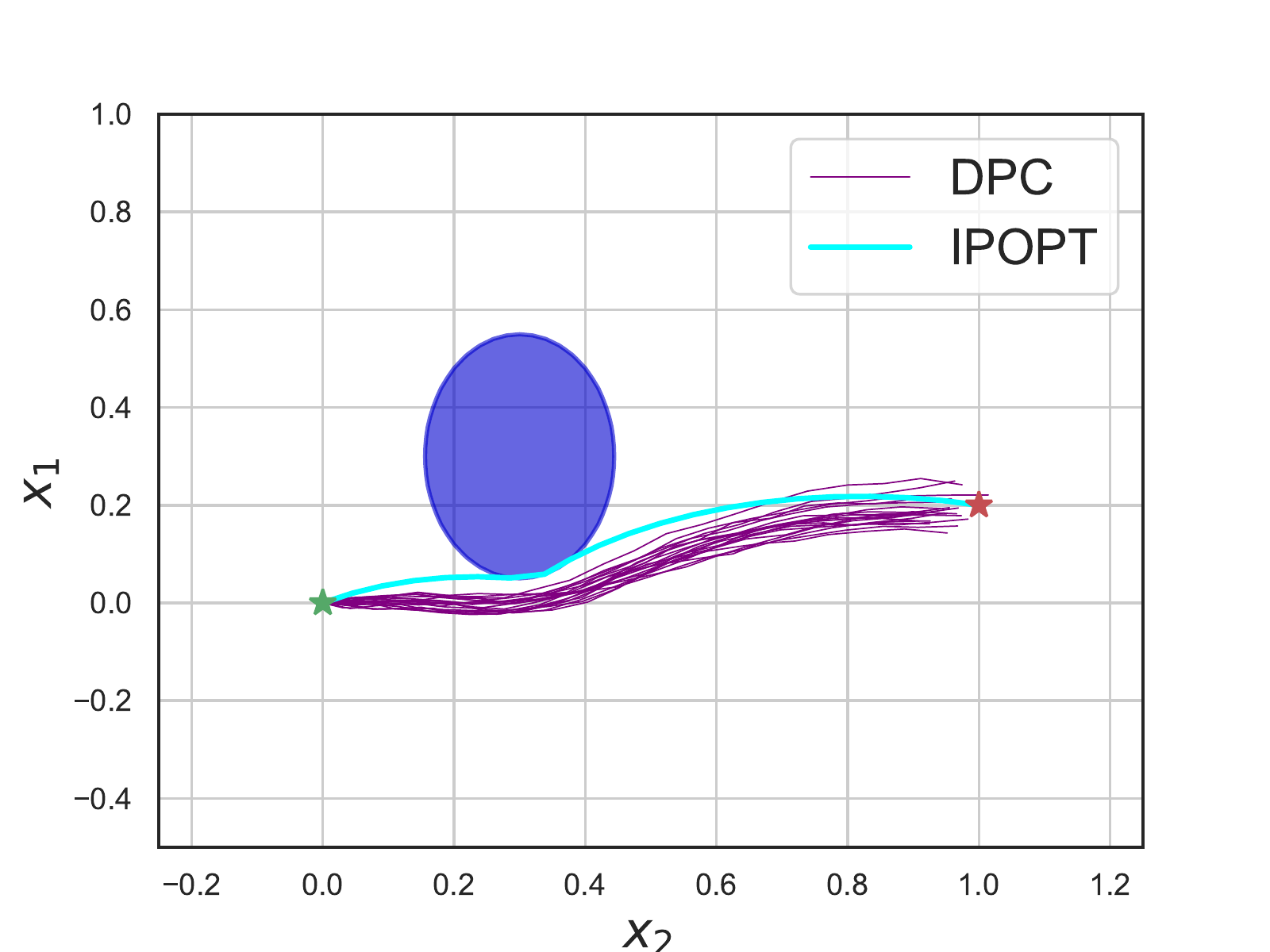}
    \includegraphics[width=0.49\linewidth]{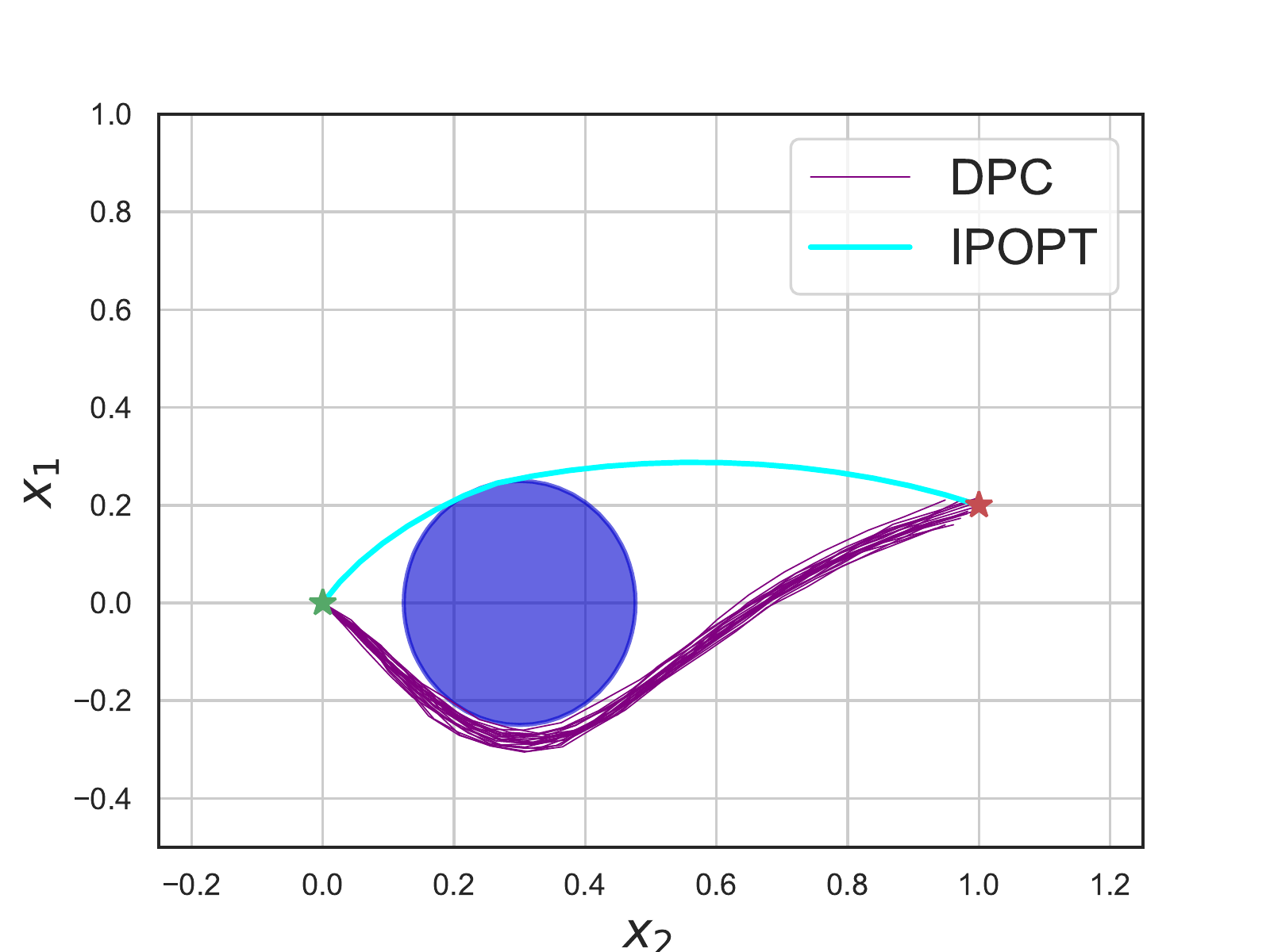}
    \includegraphics[width=0.49\linewidth]{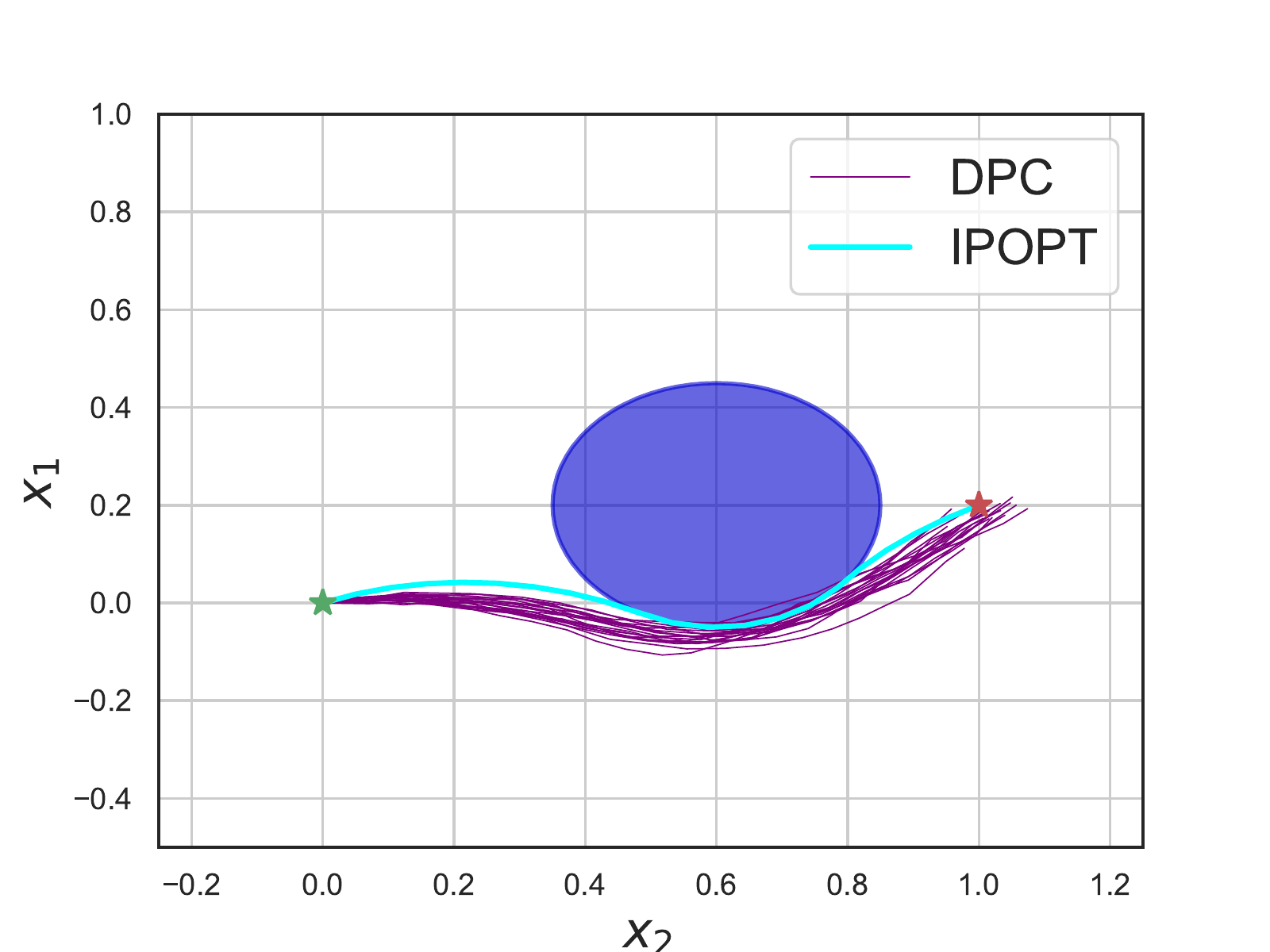}
    \caption{Examples of trajectories with different parametric scenarios of the stochastic obstacle avoidance problem~\eqref{eq:obstacle}  obtained using neural policy trained via SP-DPC Algorithm~\ref{algo:DPC_optim}. 
  Nominal nonlinear MPC  without uncertainties is computed online using IPOPT solver.}
    \label{fig:obstacle}
\end{figure}

Due to the constraint~\eqref{eq:obstacle}  the resulting parametric optimal control problem~\eqref{eq:DPC} becomes nonlinear. 
To demonstrate the computational efficiency of the proposed SP-DPC method, we evaluate the computational time of the neural policies trained via Algorithm~\ref{algo:DPC_optim} against
the deterministic nonlinear MPC implemented in CasADi framework~\citep{Andersson2019} and solved online using the IPOPT solver~\citep{Wchter2006OnTI}.
Table~\ref{tab:cpu_obstacle} shows the mean and maximum online computational time associated with the evaluation of the learned SP-DPC neural policy, compared against implicit nonlinear MPC. We show that the learned neural control policy is roughly $5$ times faster in the worst case and an order of magnitude faster on average than the deterministic NMPC solved online via IPOPT.
\add{
The  training time via Algorithm~\ref{algo:DPC_optim}
with $1000$ epochs took $17.47$  minutes.
Figure~\ref{fig:obstacle} shows stochastic trajectories generated by the proposed SP-DPC algorithm compared against nominal deterministic MPC.}
\begin{table}[htb!]
\begin{center}\caption{Online computational time of the SP-DPC neural policy, compared against implicit nonlinear MPC solved online via IPOPT.}
    \label{tab:cpu_obstacle}
\begin{tabular}{l|ll}
{online evaluation time} &   mean  [$1e^{-3}$ s] & max  [$1e^{-3}$ s]   \\ \hline
SP-DPC & 2.555  &  10.144  \\
MPC (IPOPT) & 28.362   &  53.340   \\ \hline
\end{tabular}
\end{center}
\end{table}

\section{Summary}

In this paper, we presented a learning-based stochastic parametric differentiable predictive control (SP-DPC) methodology for uncertain dynamic systems. We consider probabilistic chance constraints on the state trajectories and use a sampling-based approach encompassing variations in initial conditions, problem parameters, and disturbances. Our proposed SP-DPC policy optimization algorithm employs automatic differentiation for efficient computation of policy gradients for the constrained stochastic parametric optimization problem, which forms the basis for the differentiable programming paradigm for predictive control of uncertain systems. We provided rigorous probabilistic guarantees for the learned SP-DPC neural policies for constraint satisfaction and closed-loop system stability. Our approach enjoys better scalability than classical parametric programming solvers and is computationally more efficient than state-of-the-art nonlinear programming solvers in the online evaluation.
We substantiate these claims in three numerical examples, including stabilization of unstable systems, parametric reference tracking, and parametric obstacle avoidance with nonlinear constraints.  

\add{The proposed method assumes a known linear system dynamics model with additive uncertainties. Future work thus includes an extension to nonlinear systems. Due to the generality of the presented method, the use of nonlinear system models will not require a change in the Algorithm~\ref{algo:DPC_optim}, nor in the probabilistic guarantees given in Theorem~\ref{thm:theorem1}. The main bottleneck of the proposed method is a relatively large number of parametric and uncertainty scenarios that need to be sampled for the offline solution via Algorithm~\ref{algo:DPC_optim}. Thus future work may include adaptive sampling methods leveraging ideas of exploration and exploitation from the reinforcement learning domain. }

\begin{ack}

This research was supported by the U.S. Department of Energy, through the Office of Advanced Scientific Computing Research's “Data-Driven Decision Control for Complex Systems (DnC2S)” project. Pacific Northwest National Laboratory is operated by Battelle Memorial Institute for the U.S. Department of Energy under Contract No. DE-AC05-76RL01830. 
\end{ack}

\bibliography{ifacconf}            

\appendix

\end{document}